\documentclass{article}

\usepackage[final]{neurips_2020}

\usepackage[utf8]{inputenc} 
\usepackage[T1]{fontenc}    
\usepackage{url}            
\usepackage{booktabs}       
\usepackage{amsfonts}       
\usepackage{nicefrac}       
\usepackage{microtype}      
\usepackage{xcolor}         

\usepackage{enumitem}
\usepackage{graphicx}
\usepackage{multirow}
\usepackage{amsmath,amssymb} 
\usepackage[colorlinks,citecolor=blue,urlcolor=magenta,bookmarks=false,hypertexnames=true]{hyperref}
\usepackage{subcaption}

\usepackage[ruled, lined, linesnumbered, commentsnumbered, longend]{algorithm2e}
\DontPrintSemicolon
\usepackage{algpseudocode}

\SetCommentSty{normalfont}
\SetKwInput{KwRequire}{Require}
\SetNlSty{}{}{:}

\title{CLIP: Train Faster with Less Data}

\author{%
  Muhammad Asif Khan\thanks{Corresponding author} \\
  Qatar Mobility Innovations Center (QMIC)\\
  Qatar University\\
  Doha, Qatar \\
  \texttt{mkhan@qu.edu.qa} \\
  \And
  Ridha Hamila \\
  Department of Electrical Engineering \\
  Qatar University\\
  Doha, Qatar \\
  \texttt{hamila@qu.edu.qa} \\
  \And
  Hamid Menouar \\
  Qatar Mobility Innovations Center (QMIC) \\
  Qatar University\\
  Doha, Qatar \\
  \texttt{hamidm@qmic.com} \\ 
}

\begin{document}
\maketitle
\begin{abstract}
Deep learning models require an enormous amount of data for training. However, recently there is a shift in machine learning from model-centric to data-centric approaches. In data-centric approaches, the focus is to refine and improve the quality of the data to improve the learning performance of the models rather than redesigning model architectures. In this paper, we propose CLIP i.e., Curriculum Learning with Iterative data Pruning. CLIP combines two data-centric approaches i.e., curriculum learning and dataset pruning to improve the model learning accuracy and convergence speed. The proposed scheme applies loss-aware dataset pruning to iteratively remove the least significant samples and progressively reduces the size of the effective dataset in the curriculum learning training. Extensive experiments performed on crowd density estimation models validate the notion behind combining the two approaches by reducing the convergence time and improving generalization. To our knowledge, the idea of data pruning as an embedded process in curriculum learning is novel. 
\end{abstract}
\section{Introduction} \label{sec:intro}
Deep learning models often require a huge amount of data to train and thus to better generalize to unseen real-world examples. For instance, the state-of-the-art (SOTA) classification models e.g., VGG \cite{VGG16_ICLR2015}, ResNet \cite{ResNet_CVPR2016}, GoogleNet \cite{Inception_CVPR2015} are all trained on the ImageNet dataset \cite{imagenet_dataset} which has around 14,197,122 images (at the time of writing this paper). Although the huge amount of training data generally improves the learning accuracy, not all the training samples contribute to the training performance. In fact, it is more the quality of data that plays a major role in training. More particularly, the distribution of training data and annotation errors play can significantly affect training and generalization performance.

Recently, data-centric approaches are advocated by machine learning (ML) experts as the new paradigm for training ML models. In essence, well-crafted data techniques can help ML models to learn better and faster. We leverage two such data-centric approaches for training ML models namely curriculum learning (CL) and dataset pruning (or data pruning).

Curriculum learning is a training strategy in ML in which the training data is organized (rather than random) and exposed to the model at a predefined pace (rather than the whole data in a single training epoch). CL is inspired by human learning behavior i.e., humans learn better when using a curriculum to organize content in the order of increasing difficulty. The idea of CL was first introduced long ago in \cite{Elman1993} and formalized in \cite{CL_ICML2009}. CL has been applied in several ML tasks such as object detection and localization \cite{Tang_2018, Sangineto_PAMI2019}, machine translation \cite{Platanios_NAACL2019, Wang_ACL2019} and reinforcement learning \cite{Narvekar_JMLR2020}.

CL aims that the model to converge faster and better generalize by exposing the model to train data in a controlled manner using the difficulty scores of training samples. However, in practice, not all training samples contribute to the training. Intuitively, such samples can be eliminated from the training data. This process of eliminating the least significant training samples from a dataset is called \textit{dataset pruning}. The effectiveness of data pruning has been investigated in \cite{Li2018, Yang2022, Paul2021}.

In this paper, we propose an effective training strategy termed "CLIP" based on CL with dataset pruning. In CLIP, we start by exposing an ML model to a subset of training data and increase the training data according to a pre-defined pacing function. During the training iterations, the size of the training subset is pruned by eliminating the least-contributing samples. This effectively makes the size of the dataset smaller as the training continues. By combining CL with data pruning, the model converges faster and generalizes better.

We investigate the proposed scheme (CLIP) in the crowd counting problem which is a hot topic in computer vision \cite{Khan2022RevisitingCC}. Crowd counting is a non-trivial problem with applications in surveillance for smart policing in public places, situational awareness during disasters \cite{Sambolek2021}, traffic monitoring, and wildlife monitoring \cite{Chandana_2022}.
Traditionally crowd counting in images employed handcrafted local features such as body parts \cite{Topkaya_2014}, shapes \cite{Lin_2010}, textures \cite{Chen_2012}, edges \cite{Wu_2006}, foreground and gradients \cite{Tian_2010}. These methods perform poorly on images of dense crowds due to severe occlusions and scale variations. To overcome these challenges, CNN-based crowd counting has been introduced in \cite{CrowdCNN_CVPR2015, CrowdNet_CVPR2016}. Several state-of-the-art CNN models are developed over the course of time mainly to improve the accuracy in more challenging scenes \cite{MCNN_CVPR2016, MSCNN_ICIP2017, CSRNet_CVPR2018, TEDnet_CVPR2019, SANet_ECCV2018, SASNet_AAAI2021, MFCC_2022, SGANet_IEEEITS2022}. These SOTA crowd counting models are mainly focused on proposing model architectures with little focus on data. In CLIP, we aim to improve the performance of existing models (without any modifications to the architecture) using data-centric techniques.

The contribution of the paper is as follows:
We propose an efficient training strategy combining curriculum learning and dataset pruning termed as CLIP. CLIP effectively speeds up the model convergence and also improves model generalization to unseen data at test time. We extensively evaluated CLIP in crowd density estimation tasks using well-known crowd counting models. The results are compared against the standard training used in the original models. To our knowledge, the use of dataset pruning in curriculum learning has not been proposed earlier. Also, the use of CL and data pruning alone or together in crowd counting tasks is a new direction to improve the performance of existing crowd counting models.

\section{Related Work} \label{sec:rel_work}

The SOTA models for crowd counting and density estimation are using a convolution neural network (CNN). The first CNN-based crowd counting model was proposed in \cite{CrowdCNN_CVPR2015}. The CrowdCNN model \cite{CrowdCNN_CVPR2015} is a single-column CNN network having six convolution layers. Following the approach, there has been a long list of CNN-based model of different types including multi-column networks \cite{MCNN_CVPR2016, CrowdNet_CVPR2016, SCNN_CVPR2017, CMTL_AVSS2017}, pyramid networks \cite{MSCNN_ICIP2017, SANet_ECCV2018, SGANet_IEEEITS2022}, encoder-decoder models \cite{TEDnet_CVPR2019, MobileCount_PRCV2019}, and transfer learning based models \cite{CSRNet_CVPR2018, CANNet_CVPR2019, GSP_CVPR2019, TAFNET_2022}. The best-performing models today are those using transfer learning which uses a pretrained image classification model such as VGG \cite{VGG16_ICLR2015}, ResNet \cite{ResNet_CVPR2016} or Inception \cite{Inception_CVPR2015} as a front-end to extract features and then a small CNN network uses these features to estimate the crowd density. However, typically these models are very large requiring a long time to train and converge.

Curriculum learning has not been applied explicitly in crowd counting tasks. However, it has been evaluated in other computer vision tasks. A notable work on CL in \cite{Guy_2019} investigated the potential benefits of CL in the classification problem using CIFAR10 and CIFAR100 datasets. Their results show that CL brings significant accuracy improvement. The work also shows that different pacing functions converge to similar final performances. A recent study on the CL \cite{Wu2021} with an extensive set of experiments shows that CL significantly reduces the convergence time while improving the learning performance. The results also show that it is actually the pacing function rather than curricula (alone) that improves learning performance. 

The impact of data pruning and its potential benefits are investigated in \cite{Yang2022, Paul2021}. In \cite{Yang2022}, authors propose data pruning under generalization constraint. In this method, the redundant training samples are identified and removed based on the loss function. The model accuracy using the pruned dataset (with fewer samples than the original dataset) is comparable (slightly less) to the model trained on the whole dataset. In \cite{Paul2021}, authors used two types of scores (gradient normed and L2-norm) to identify the least significant samples during the first few epochs of the training phase and prune them to get a compact dataset for the remaining training phase. The method shows that better accuracy was achieved on the CIFAR10 dataset even after removing half of the training samples.

\section{Proposed Scheme} \label{sec:scheme}

The proposed scheme i.e., is implemented by combining (i) curriculum learning and (ii) data pruning. 
\par
\begin{figure}
\centering
\subcaptionbox{Linear pacing function (Solid lines represent full data and broken lines represent pruned data).}
{\includegraphics[width=0.45\columnwidth]{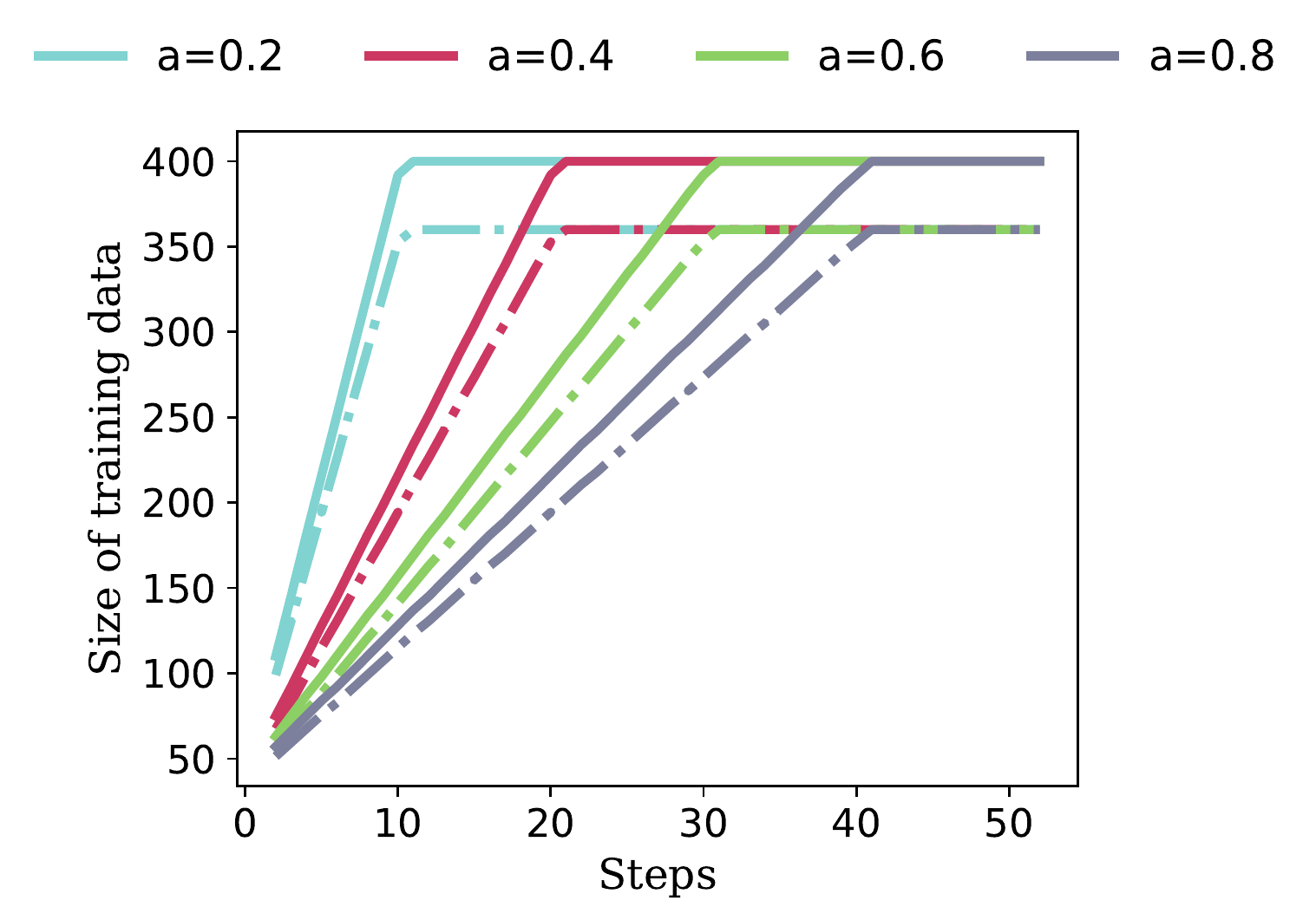}}
\hspace{1em}
\subcaptionbox{Quadratic pacing function (Solid lines represent full data and broken lines represent pruned data).}
{\includegraphics[width=0.45\columnwidth]{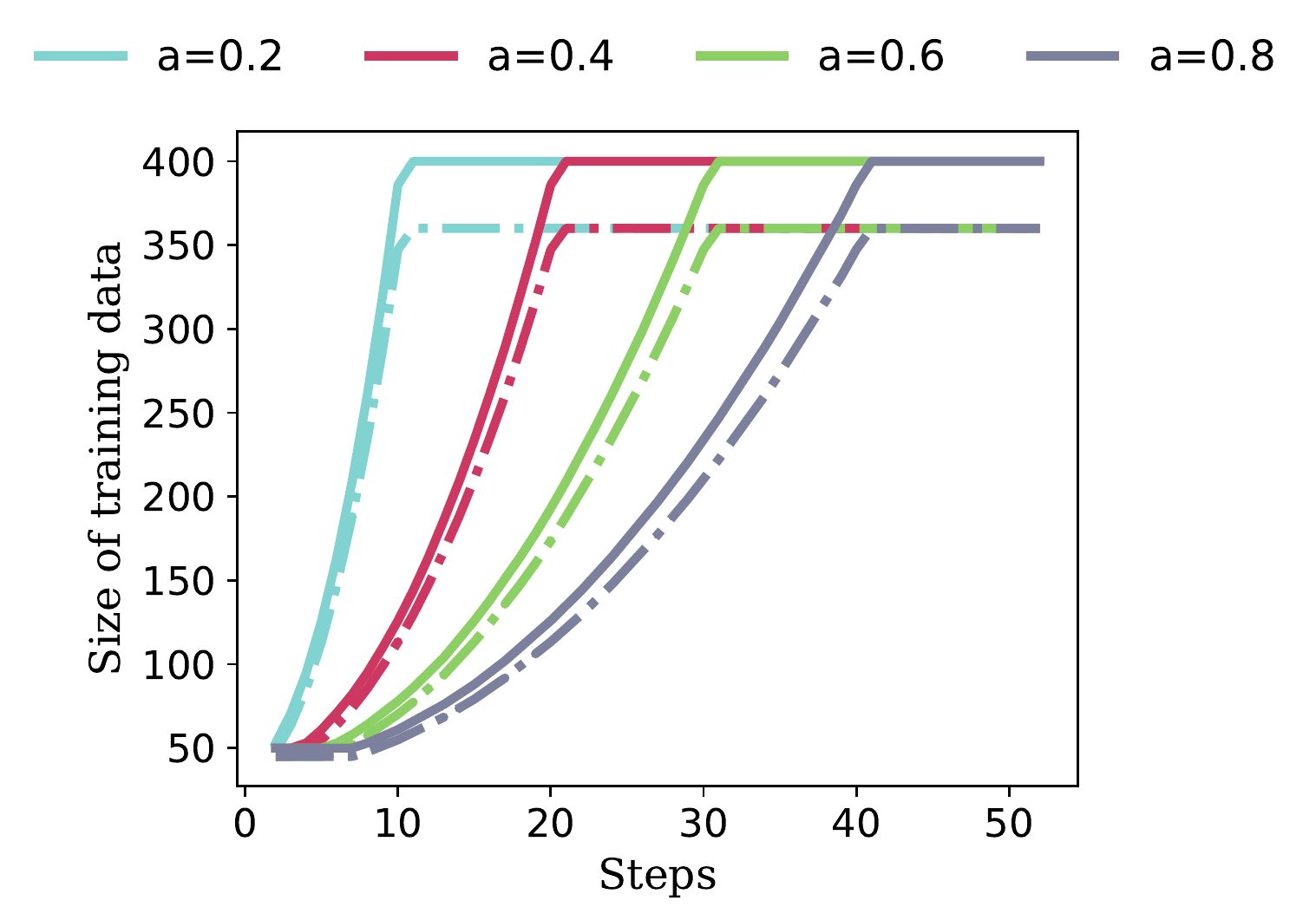}}
\caption{Illustration of two pacing functions in curriculum learning with dataset pruning ($\epsilon = 0.05$).}
\label{fig:pacing_functions}
\end{figure}

Curriculum learning has two components i.e., a scoring function and a pacing function. A pre-training model is used to calculate per-sample loss that serves as the sample scores. We used a CSRNet \cite{CSRNet_CVPR2018} to calculate per-sample losses that serve as sample scores (difficulties). A pacing function is then used to initially expose the model to a small subset of data and the size of the subset is iteratively increased. We used two different pacing functions (linear and quadratic) (Fig. \ref{fig:pacing_functions}) in our study however, due to very close results, we are reporting results for the latter function only.
\par
\begin{figure*}
\centering
\subcaptionbox{Mall dataset}
{\includegraphics[width=0.3\textwidth]{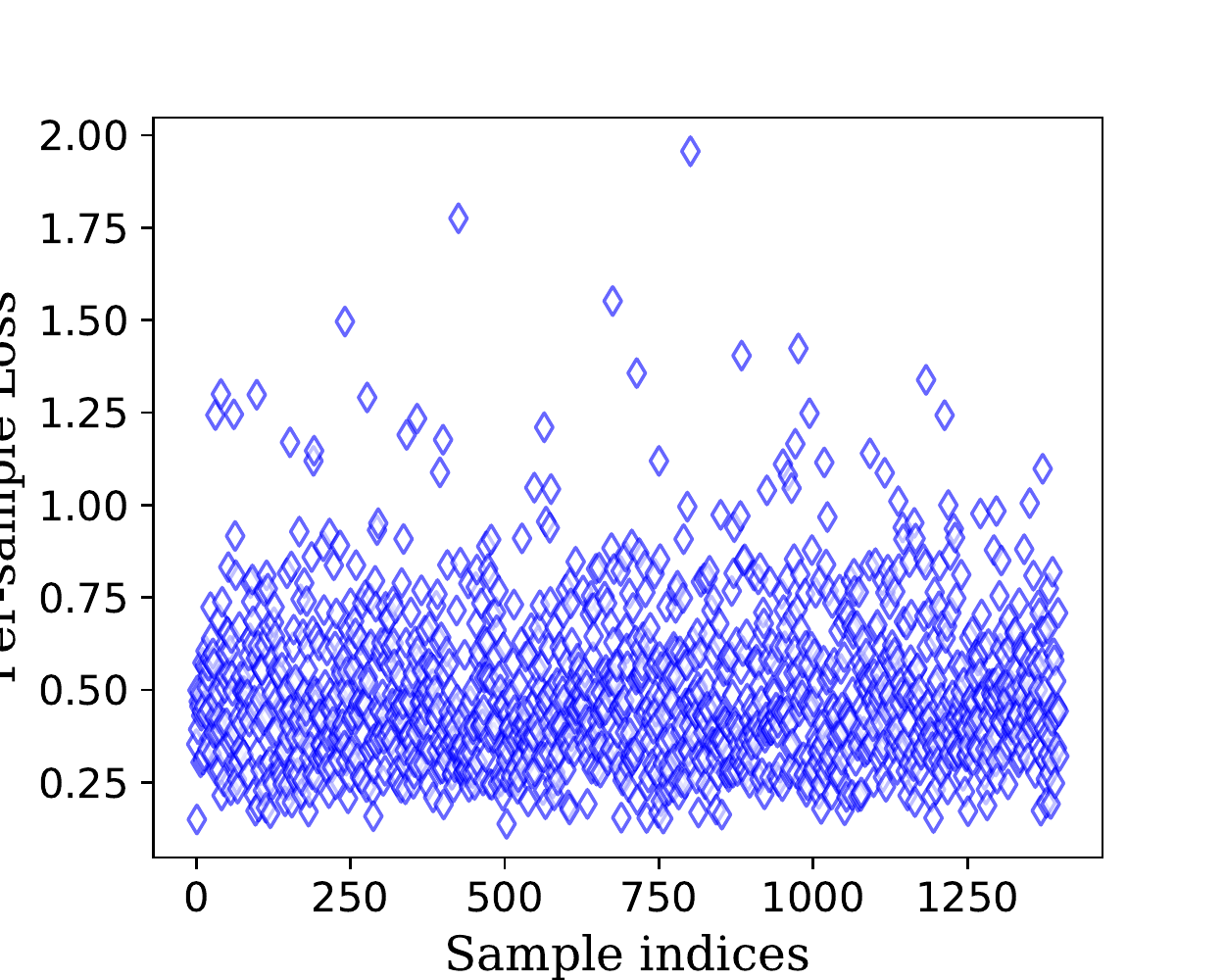}} \hspace{1em}
\subcaptionbox{ShanghaiTech Part-A}
{\includegraphics[width=0.3\textwidth]{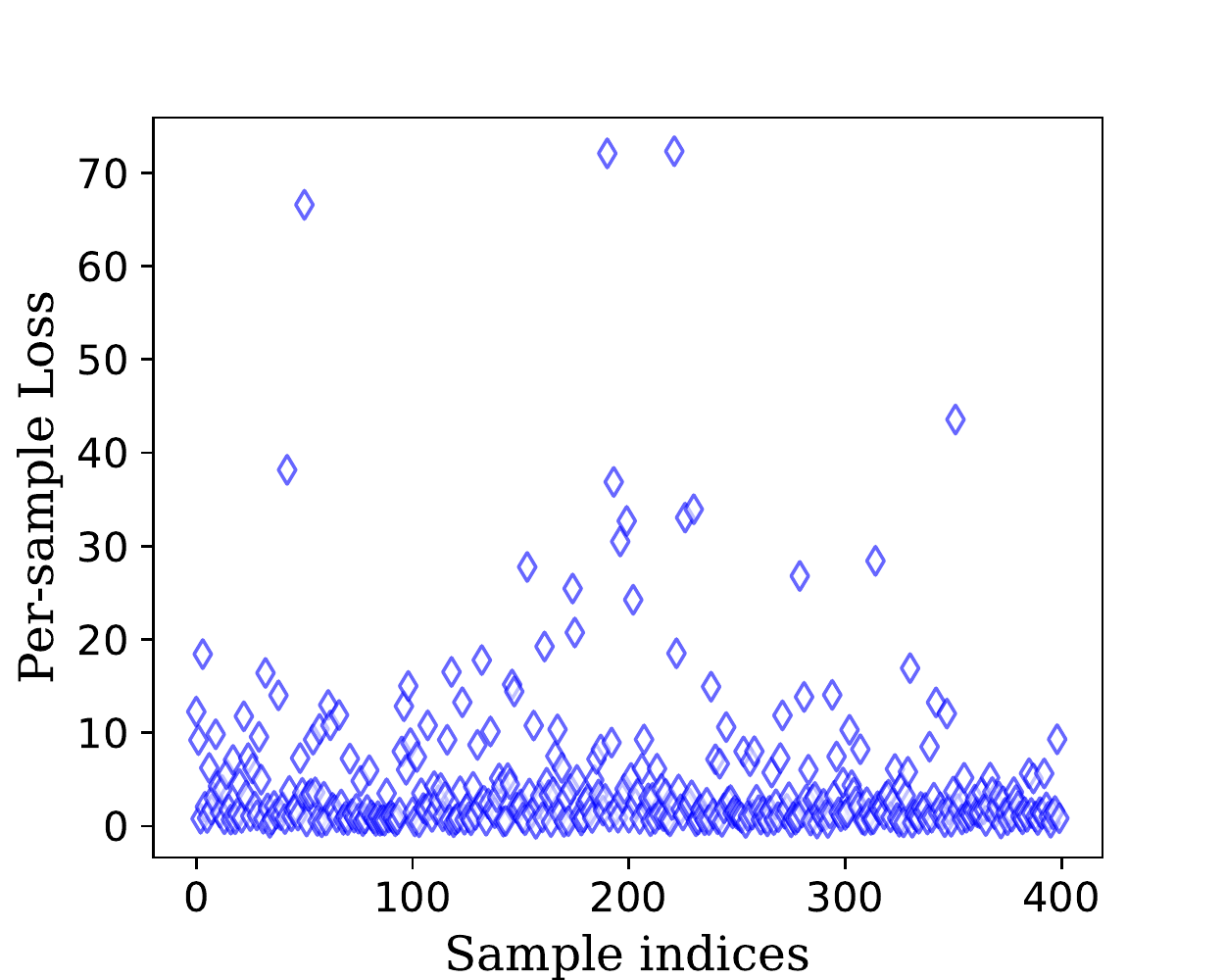}} \hspace{1em}
\subcaptionbox{ShanghaiTech Part-B}
{\includegraphics[width=0.3\textwidth]{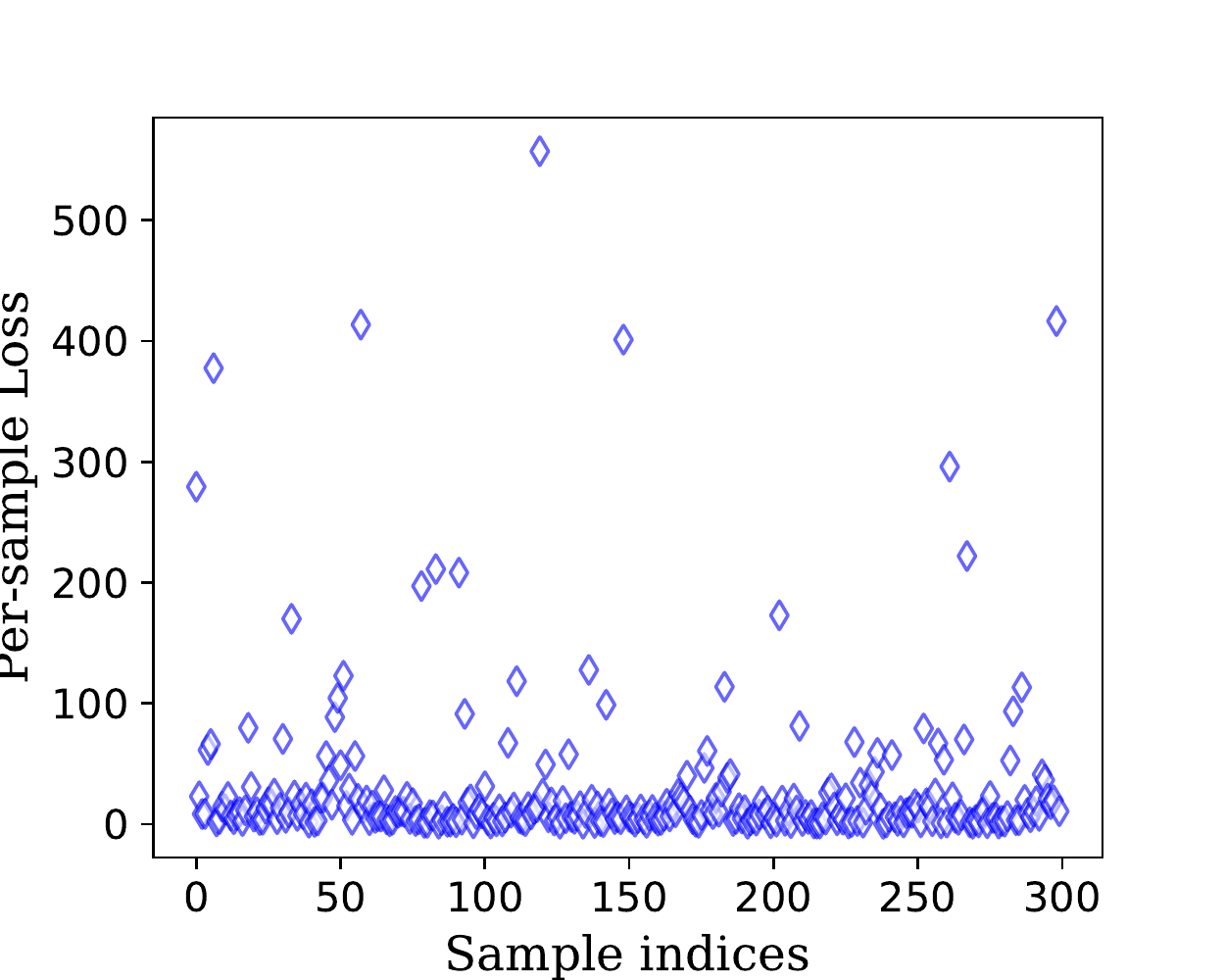}} 
\\\vspace{1em}
\caption{Per-sample losses of ShanghaiTech part-A (top) and ShanghaiTech Part B (bottom) dataset using various models.}
\label{fig:per_sample_losses}
\end{figure*}

The proposed scheme (CLIP) additionally applies data pruning i.e., removing the least significant samples from the training subsets during the training. To implement data pruning in CLIP, we first empirically found the loss value of each sample of the original training dataset which showed that different samples contribute differently in the learning process with some samples having very little contribution. Thus, it is intuitive that such samples can be eliminated without significantly impacting the model's average training loss. The per-sample losses of different datasets can be observed in Fig. \ref{fig:per_sample_losses}.
\par
To integrate data pruning in the curriculum learning process, we define additional parameters $\epsilon$. The value of $epsilon$ can be either fixed or can be calculated using an increasing function with decreasing rate (or vice versa depending on the size of the dataset). Algorithm \ref{algo:clip} illustrates the CLIP procedure.

\begin{algorithm}
\caption{CLIP - Curriculum Learning with Iterative dataset Pruning.} \label{algo:clip}
\KwRequire{scoring function $f$, pacing function $g$, data $X$, sample elimination ratio $\epsilon$}
\KwResult{mini-batches [$B_1, B_2, ... B_M$]}

$results$ = sort $X$ using $f$ \;
\For{$i=1, \cdots M$}{
    $size \leftarrow g(i)$ \;
    $size = size - \epsilon (size)$ \;
    $X_i = X[1, ..., size]$ \;
    uniformly sample $B_i$ to $results$ \;
    append $B_i$ to $result$ \;
    }
\KwRet{$results$}
\end{algorithm}

\subsection{Model Training}
We choose three models i.e., MCNN \cite{MCNN_CVPR2016}, CSRNet \cite{CSRNet_CVPR2018} and CSRNet\_lite \cite{LCDnet_2022} and three different datasets i.e., ShanghaiTech Part-A and Part-B, Mall \cite{Mall_dataset2012}, and CARPK \cite{CARPK_dataset}.
The original labels are in the form of dot annotations which are used to create density maps that serve as ground truth for the images. A density map is generated by convolving a delta function $\delta(x - x_i)$ with a Gaussian kernel $G_\sigma$, where $x_i$ is a pixel containing the head position.
\begin{equation}
    D = \sum_{i=1}^{N}{ \delta(x-x_i) * G_\sigma}
\end{equation}
where, $N$ denotes the total number of annotated points (i.e., the total count of heads) in the image. We empirically determined a fixed value of $\sigma$ that provides a good estimation of the head sizes for each individual dataset. To prevent overfitting, we use standard data augmentation techniques such as horizontal flipping, and random brightness and contrast. We use Adam optimizer \cite{Adam_ICLR2015} with a learning rate of $0.0001$. The loss function used is the standard Euclidean distance between the target and predicted density maps which are defined in Eq. \ref{eq:mse}.

\begin{equation} \label{eq:mse}
    L(\Theta) = \frac{1}{N} \sum_{1}^{N}{ ||D(X_i;\Theta) - D_i^{gt}||_2^2}
\end{equation}

where $N$ is the number of samples in training data, $D(X_i;\Theta)$ is the predicted density map with parameters $\Theta$ for the input image $X_i$, and $D_i^{gt}$ is the ground truth density map.

To implement data pruning, we used a fixed small value of $\epsilon = 0.05$ such that it eliminates a small number of least significant samples from the training subset in each iteration until the training data size reaches the maximum (original dataset size - total samples removed).

\section{Evaluation and Results} \label{sec:results}

We first investigated the effectiveness of the proposed scheme. The loss curve in Fig. \ref{fig:loss} shows a clear advantage of CLIP over standard training. The training loss drops very quickly in CLIP with much fewer samples fed to the model as compared to the standard training. This implies the correctness of our intuition which is further investigated with an extensive set of experiments on several datasets.

\subsection{Evaluation Metrics} \label{subsec:metrics}
We evaluated the performance of CLIP using four widely used metrics i.e., Mean Absolute Error (MAE) and Grid Average Mean Error (GAME), Structural Similarity Index (SSIM), and Peak Signal-to-Noise Ratio (PSNR). The first two metrics evaluate the counting accuracy of the model whereas the latter two evaluate the quality of the predicted density maps.

MAE and GAME can be calculated using the following Eq. \ref{eq:mae} and Eq. \ref{eq:game}:

\begin{equation} \label{eq:mae}
    MAE = \frac{1}{N} \sum_{1}^{N}{(e_n - \hat{g_n})}
\end{equation}

where, $N$ is the size of the dataset, $g_n$ is the target or label (actual count) and ${e_n}$ is the prediction (estimated count) in the $n^{th}$ crowd image.

\begin{equation} \label{eq:game}
    GAME = \frac{1}{N} \sum_{n=1}^{N}{ ( \sum_{l=1}^{4^L}{|e_n^l - g_n^l|)}}
\end{equation}
The value of $L=4$ denoted that each density map is divided into a $4\times4$ grid size with a total of $16$ patches.

The SSIM and PSNR metrics can be calculated using the below Eq. \ref{eq:ssim} and Eq. \ref{eq:psnr}:

\begin{equation} \label{eq:ssim}
    SSIM (x,y) = \frac{(2\mu_x \mu_y + C_1)  (2\sigma_x \sigma_y C_2)}  {(\mu_z^2 \mu_y^2 + C_1)  (\mu_z^2 \mu_y^2 + C_2)}
\end{equation}
where $\mu_x, \mu_y, \sigma_x, \sigma_y$ denotes the means and standard deviations of the labels (density maps) and predictions (density maps), respectively.

\begin{equation} \label{eq:psnr}
    PSNR = 10 log_{10}\left( \frac{Max(I^2)}{MSE}  \right)
\end{equation}
where $Max(I^2)$ is the maximal pixel value in the image, e.g., 255 if pixel values are stored as 8-bit unsigned integer type.


\begin{figure}
\centering
\includegraphics[width=0.5\columnwidth]{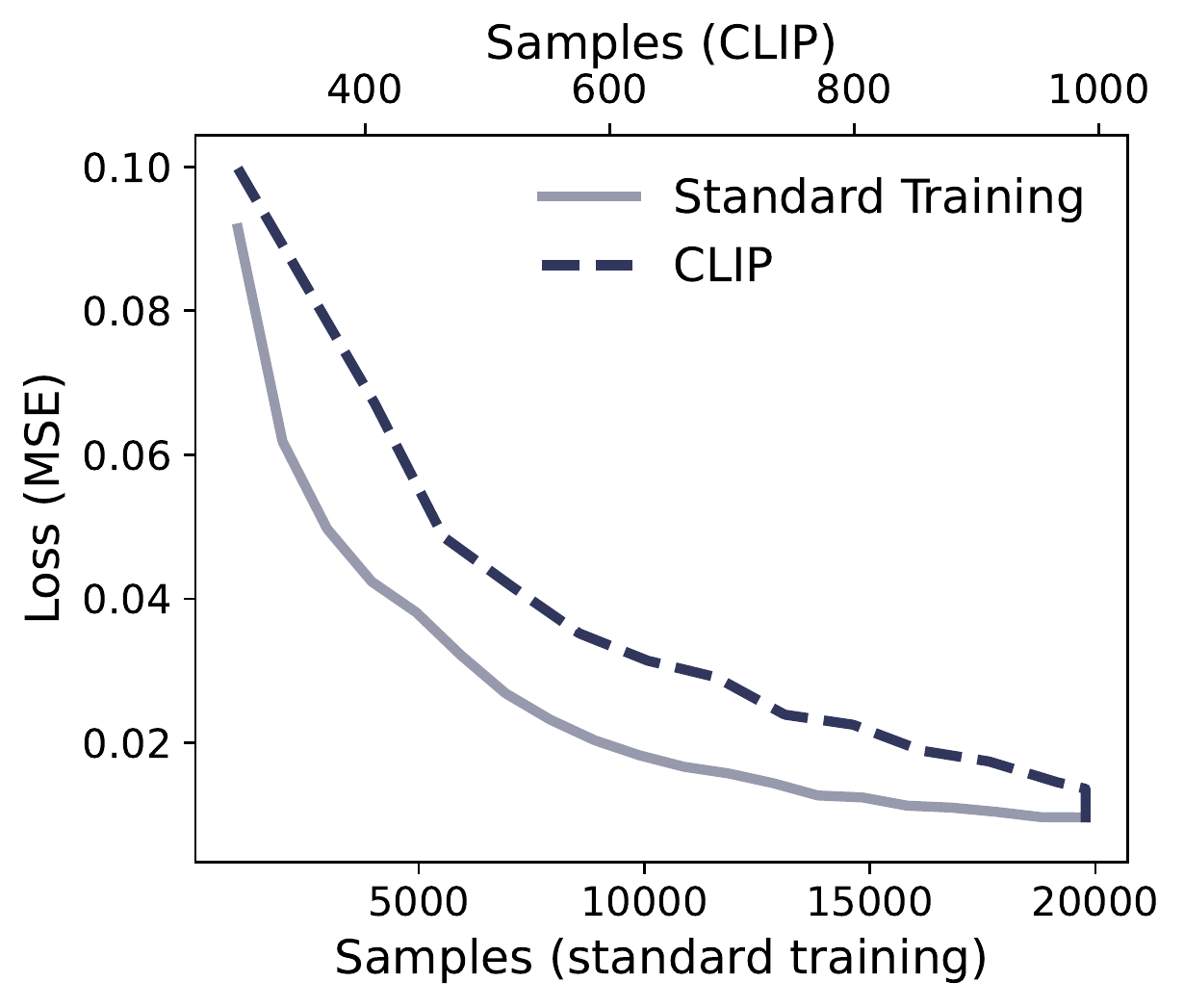}
\caption{Loss convergence comparison. The bottom X-axis shows the number of samples used in standard training. The top X-axis shows the number of samples used in CLIP. The number of samples increases from left to right. The model loss decreases from left to right.}
\label{fig:loss}
\end{figure}

\begin{figure}
\centering
\includegraphics[width=0.8\columnwidth]{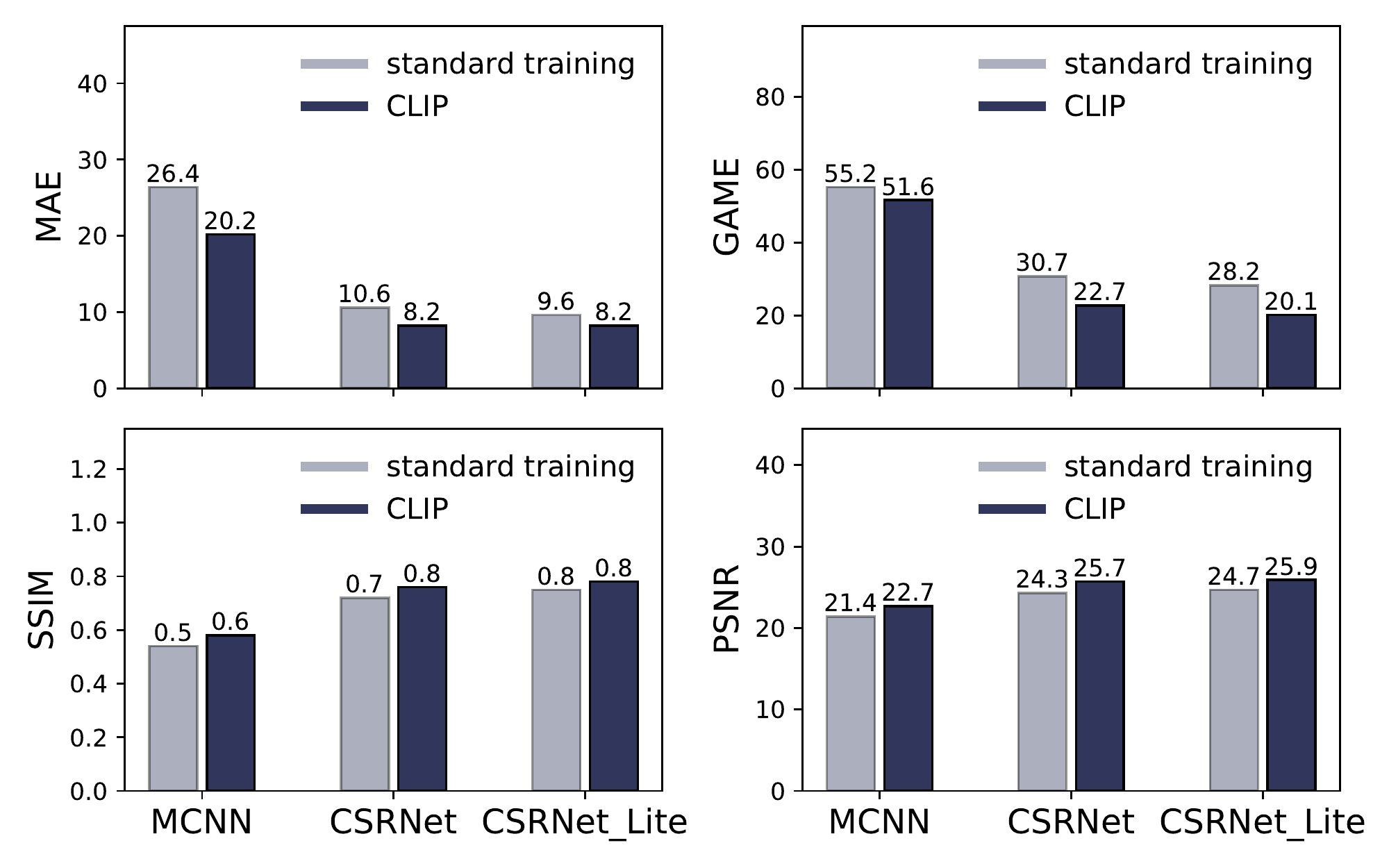}
\caption{Performance gain achieved using CLIP versus standard training on ShanghaiTech Part-B \cite{MCNN_CVPR2016} dataset.}
\label{fig:accuracy_B}
\end{figure}

\begin{figure}
\centering
\includegraphics[width=0.8\columnwidth]{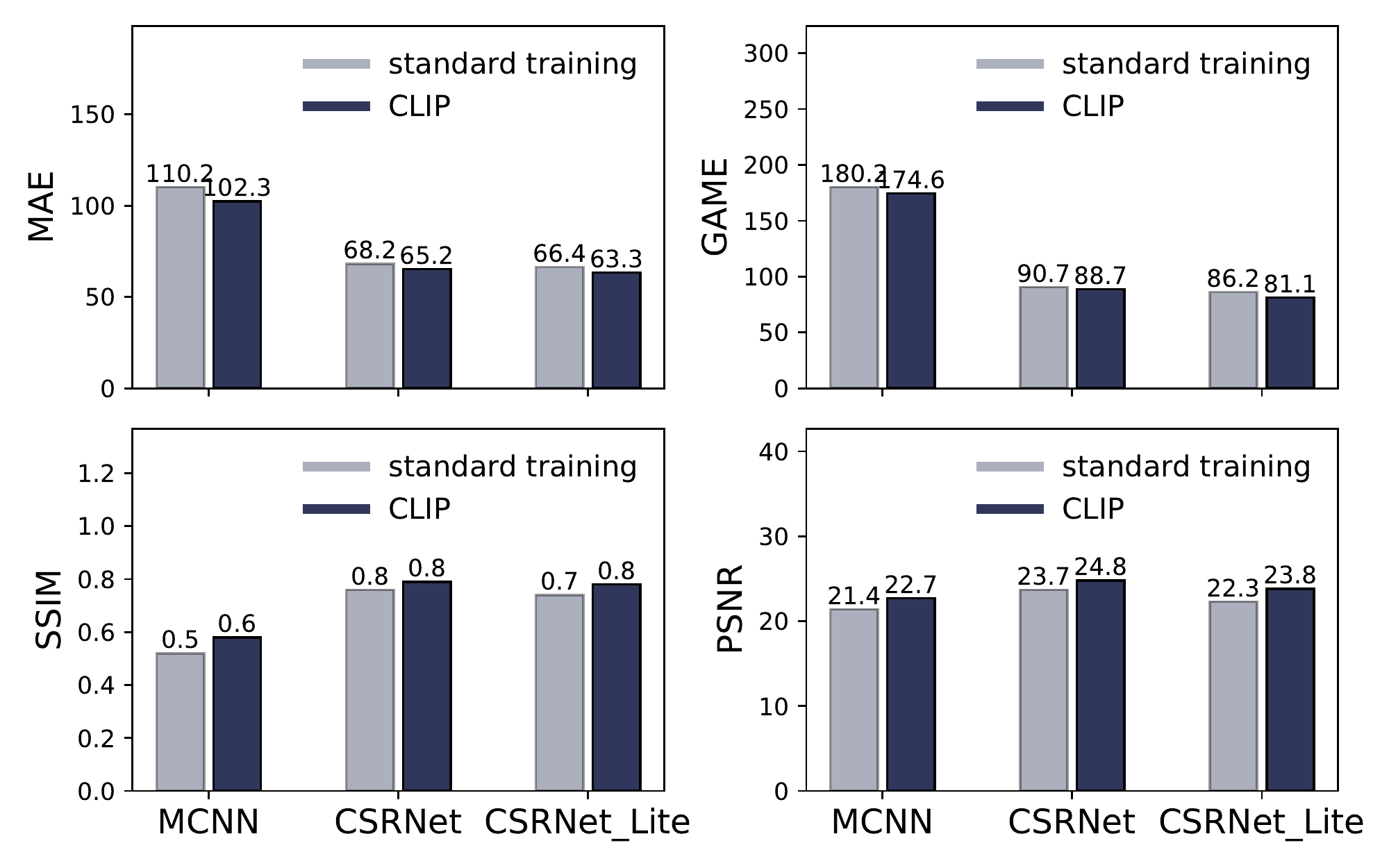}
\caption{Performance gain achieved using CLIP versus standard training on ShanghaiTech Part-B \cite{MCNN_CVPR2016} dataset.}
\label{fig:accuracy_A}
\end{figure}

The results for all the four metrics on ShanghaiTech Part-B and ShanghaiTech Part-A datasets are presented in Fig. \ref{fig:accuracy_B} and \ref{fig:accuracy_A} showing improvement using CLIP as follows: On ShanghaiTech Part-B, CLIP reduces the MAE (as compared to standard training) by $26.4 \Rightarrow 20.2 \; (23\% \; \text{reduction})$ using MCNN \cite{MCNN_CVPR2016}, $10.6 \Rightarrow 8.2 \; (22\% \; \text{reduction})$ using CSRNet \cite{CSRNet_CVPR2018}, and $9.6 \Rightarrow 8.2 \; (14\% \; \text{reduction})$ using CSRNet\_lite \cite{LCDnet_2022}. On ShanghaiTech Part-A, CLIP reduces the MAE by $110.2 \Rightarrow 102.3 \; (7\% \; \text{reduction})$ using MCNN \cite{MCNN_CVPR2016}, $68.2 \Rightarrow 65.2 \; (4\% \; \text{reduction})$ using CSRNet \cite{CSRNet_CVPR2018}, and $66.4 \Rightarrow 63.3 \; (4\% \; \text{reduction})$ using CSRNet\_lite. A similar performance was achieved using the GAME metric as well. Although there is a clear gain in accuracy over both datasets using three different models, the actual advantage of CLIP over standard training is achieving higher accuracy with faster training and less number of training samples as depicted in Fig. \ref{fig:loss}. The reader is encouraged to notice the minor improvements over SSIM and PSNR values in Fig. \ref{fig:accuracy_B} and \ref{fig:accuracy_A} which represent the quality of the predicted density maps.

\begin{figure*}
\centering
\includegraphics[width=0.8\textwidth]{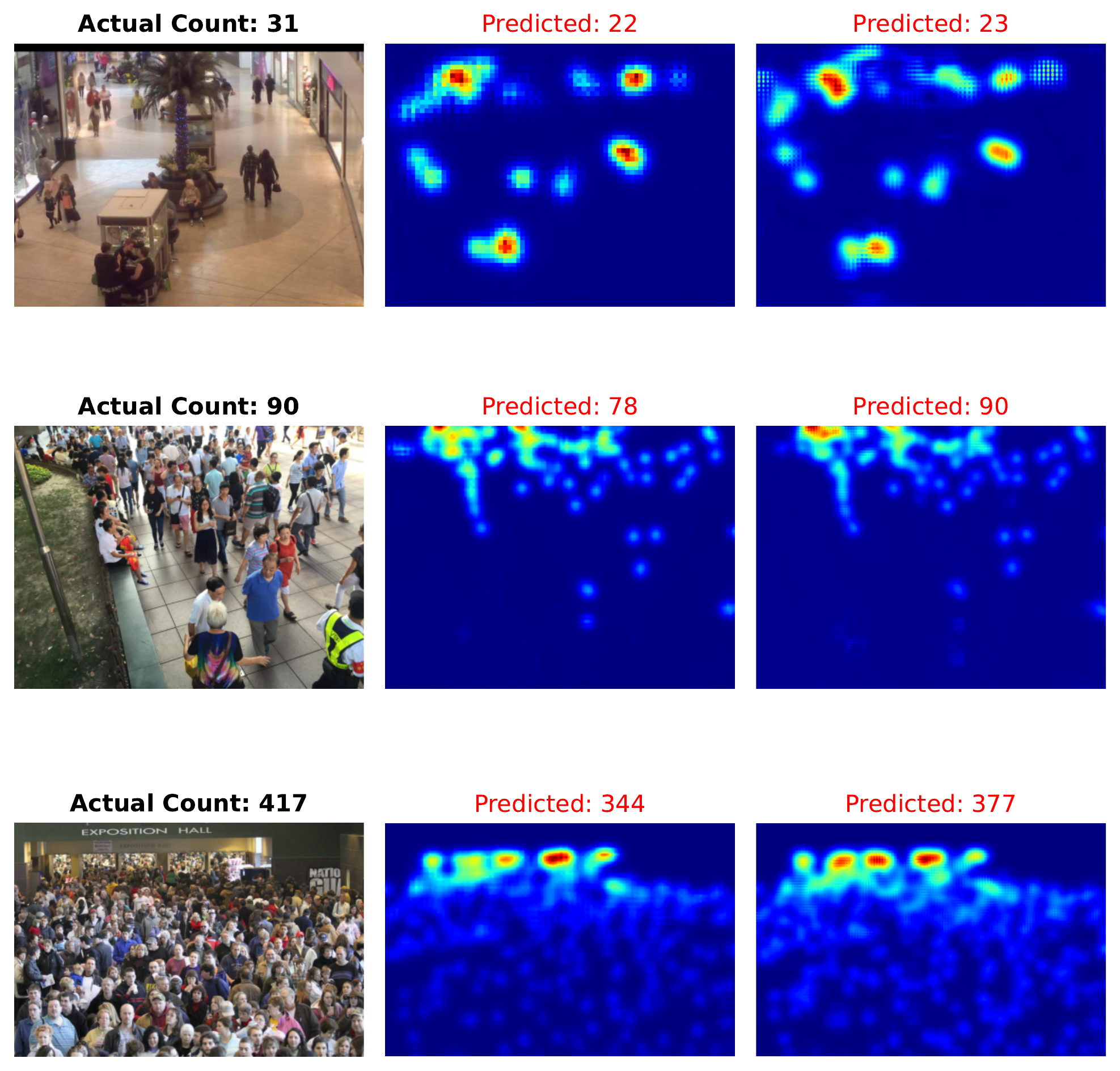}
\caption{Sample predictions over three datasets. The left column shows sample images each from Mall \cite{Mall_dataset2012}, ShanghaiTech Part-B \cite{MCNN_CVPR2016}, and ShanghaiTech Part-A datasets \cite{MCNN_CVPR2016}. The middle column shows predictions using CSRNet \cite{CSRNet_CVPR2018} model with standard training. The right column shows predictions using CSRNet with CLIP training.}
\label{fig:predictions}
\end{figure*}

We also evaluated the performance of the proposed scheme (CLIP) over the Mall dataset \cite{Mall_dataset2012} and CARPK dataset \cite{CARPK_dataset} and found similar improvements of CLIP compared to standard training. Some interesting predictions showing better predictions using CLIP as compared to standard training are shown in Fig. \ref{fig:predictions}.

\section{Conclusion} \label{sec:conclusion}
In this paper, we propose CLIP - curriculum learning with iterative dataset pruning. CLIP is an efficient training strategy improving model learning performance as well as reducing convergence time. The evaluation results validate the benefits of CLIP in crowd counting tasks. We believe CLIP can be effectively applied in other deep learning tasks particularly when the dataset size is too large. CLIP can be beneficial when the original dataset has erroneous samples with noisy labels.

\bibliographystyle{plainnat}
\bibliography{biblio}







\end{document}